\documentclass[runningheads]{llncs}
\usepackage[T1]{fontenc}
\usepackage{graphicx}
\graphicspath{ {./images/} }
\usepackage{xcolor}
\usepackage{soul}
\usepackage{amsmath}
\usepackage{amssymb}
\usepackage{amsfonts}
\usepackage{url}
\usepackage{multirow}
\usepackage{orcidlink}
\usepackage{rotating}
\usepackage{lscape}
\usepackage{comment}
\usepackage[graphicx]{realboxes}

\begin{document}

\title{Adaptive Class Learning to Screen Diabetic Disorders in Fundus Images of Eye\thanks{This research was supported by the J. C. Bose National Fellowship, grant no. JCB/2020/000033 of S. Mitra.}}
\author{Shramana Dey\inst{1}\orcidID{0009-0008-9815-1558} \and
Pallabi Dutta\inst{1}\orcidID{0000-0002-5231-4023} \and
Riddhasree Bhattacharyya\inst{1}\orcidID{0009-0007-9824-2874} \and Surochita Pal\inst{1}\orcidID{0000-0002-6749-4426} \and Sushmita Mitra\inst{1}\orcidID{0000-0001-9285-1117} \and Rajiv Raman\inst{2}\orcidID{0000-0001-5842-0233}}

\authorrunning{S. Dey et al.}

\institute{Machine Intelligence Unit, Indian Statistical Institute, \\ 203, B. T. Road, Kolkata - 700108, West Bengal, India \and Shri Bhagwan Mahavir Vitreoretinal Services, Sankara Nethralaya, Chennai, India}
\maketitle              

\begin{abstract}
The prevalence of ocular illnesses is growing globally, presenting a substantial public health challenge. Early detection and timely intervention are crucial for averting visual impairment and enhancing patient prognosis. This research introduces a new framework called \textit{Class Extension with Limited Data (CELD)} to train a classifier to categorize retinal fundus images. The classifier is initially trained to identify relevant features concerning \textit {Healthy} and \textit {Diabetic Retinopathy (DR)} classes and later fine-tuned to adapt to the task of classifying the input images into three classes, \textit {viz. Healthy, DR and Glaucoma}. This strategy allows the model to gradually enhance its classification capabilities, which is beneficial in situations where there are only a limited number of labeled datasets available. Perturbation methods are also used to identify the input image characteristics responsible for influencing the model's decision-making process. We achieve an overall accuracy of 91\% on publicly available datasets.

\end{abstract}
\section{Introduction}
With the rapid growth in the global population, the number of individuals diagnosed with diabetes is increasing at an alarming rate. Diabetes, a metabolic disorder characterized by high blood sugar levels, often leads to various visual impairments. Diabetic individuals must undergo regular eye screening due to the strong correlation between diabetes and eye abnormalities. A significant challenge lies in the shortage of trained eye care professionals, which hampers effective screening and treatment \cite{basu2021segmentation}. Diabetes is a precursor for several vision-threatening diseases, notably diabetic retinopathy (DR) and Glaucoma \cite{bourne2013causes}. Approximately one-third of diabetics are likely to develop DR, and those with diabetes are twice as likely to be afflicted by Glaucoma as compared to non-diabetic individuals \cite{morya2023diabetes}. Both DR and Glaucoma often progress silently until significant vision loss occurs, potentially leading to irreversible blindness. 

Regular, automated, non-invasive screening is crucial for early detection. This helps prevent the progression of these diseases in the preliminary stage, especially in remote regions with limited access to trained professionals. These technologies can help identify at-risk individuals early, allowing timely intervention \cite{fatima2024enhancing}. Deep learning (DL) \cite{lecun2015deep} is a popular choice in smart healthcare, for automating the screening process. Feature extraction with minimal human assistance, scalability, and high output efficiency, are some of the key factors attributed to the acceptability of deep learning methodologies in healthcare. This helps make the screening process efficient in terms of time, while coping with the limited number of trained professionals. 

Several studies have utilized deep learning models for DR classification. For instance, the multi-resolution convolutional attention network (MuR-CAN) \cite{madarapu2024multi} emphasizes discriminative features using a multi-dilation attention block, with depth-wise convolution layers at various dilation rates to capture multi-scale spatial information. The DRNet13  \cite{shamrat2024advanced} has been developed for automated DR stage classification. The Modified Generative Adversarial-based Crossover Salp Grasshopper (MGA-CSG) \cite{navaneethan2024enhancing} approach predicts and classifies diabetic retinal diseases using fundus image datasets.
Research has also focused on analyzing multiple pathologies from eye fundus images \cite{grover2023detection}, using various CNN-based models like LeNet, AlexNet, Inception, VGG, and ResNet, for diagnosing Glaucoma and DR. Vision transformers \cite{dosovitskiy2020image} have also been employed for ocular disease detection and classification using fundus images.

Deep learning models require large volumes of annotated data to learn features for accurate prediction. Given the limited number of trained professionals, this becomes a major challenge. Consequently, automated detection of DR and Glaucoma, from eye fundus images, gets constrained by data scarcity issues. Most publicly available datasets provide retinal images without detailed labeling of the affected region(s). Transfer learning has been employed \cite{basu2020deep} to handle the scarcity of data by adapting weights from models trained on larger datasets. However, this often faces challenges like catastrophic forgetting and degraded performance due to domain shifts. 

This paper proposes a framework - Class Extension with Limited Data (CELD), which trains classifiers to recognize additional (new) classes over time without forgetting relevant features from the previously learned classes. This framework is particularly useful in scenarios where new data classes get gradually incorporated. In real-life scenarios, as new and rarer ocular diseases are discovered or become more prevalent, it becomes necessary to update the diagnostic model to recognize these new conditions while still retaining the ability to diagnose previously known diseases. The CELD framework addresses the data scarcity and imbalance issues prevalent in DR and Glaucoma classification when compared to healthy samples. 
Unlike transfer learning, which requires a substantial volume of data to fine-tune the model, the proposed framework updates the model as new data becomes available; without having to retrain from scratch. This approach allows the network to continuously learn from smaller, progressive batches; thereby, making it resource-efficient and scalable for dynamic environments. Several controlled data-perturbation techniques are incorporated to analyze the decision-making process of our model. This adds explainability to address the significance of each input attribute towards model behavior. The key contributions of the research are listed below.

\begin{itemize}
    \item The class adaptation in CELD progresses incrementally. A deep neural network is first trained to classify fundus images into healthy and DR classes. It is then extended to additionally  classify Glaucoma, to transform to a three-class learning model. 
    \item The CELD framework prevents catastrophic forgetting of previous learning, while leveraging existing knowledge to learn new classes in the presence of limited data. 
    \item Detailed empirical study and analysis of the CELD framework establishes its robustness to data and use of fewer computational resources.
    \item Feature relevance is explored, through data perturbation, to analytically observe changes in model performance. 
    \end{itemize}  

The remaining sections of the paper are organized as follows. Section \ref{meth} describes the CELD framework. Section \ref{res} outlines the experimental results to study the performance of our proposed framework CELD. Finally Section \ref{concl} concludes the article. 

\section{Methodology} 
\label{meth}

\begin{figure}[!ht]
    \centering
    \includegraphics[width=\linewidth]{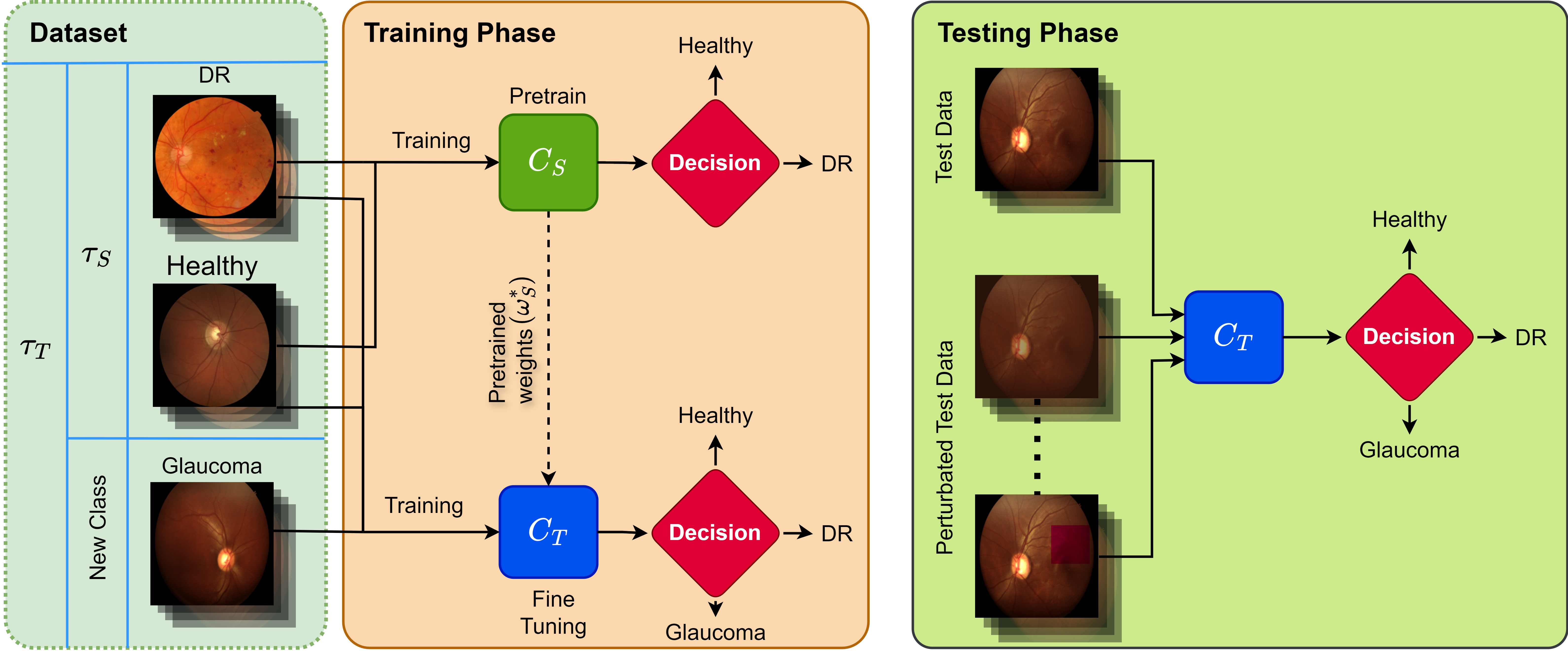}
    \caption{The basic workflow of the proposed CELD-Framework}
    \label{fig:workflow}
\end{figure}

A significant challenge in the task of retinal image classification is the limited availability of annotated data, which constrains the generalizability of the model. Specifically, there is a disproportionate ratio of healthy and DR data, with Glaucoma data being even scarcer. This imbalance complicates the task of improving classification accuracy. While data augmentation might intuitively address this issue, it risks overfitting, resulting in an inefficient model \cite{garcea2023data}. To effectively manage this data imbalance, the proposed CELD framework helps to classify fundus images as healthy, DR-affected, or Glaucoma-affected. Additionally, we evaluate the model's decision-making process using explainability methods based on perturbation techniques. A schematic workflow is provided in Fig. \ref{fig:workflow}.

The objective of this study is to develop a detection system for retinal color fundus images of three classes: healthy, DR, and Glaucoma. The subsequent parts of this section provide a detailed explanation of the classifier architecture, the proposed CELD framework, and the perturbation-based explainable methods used to gain insights regarding the decision-making process of the model.

\subsection{Class Extension with Limited Data (CELD)}
A deep learning model tends to experience "catastrophic forgetting", losing previously learned patterns when trained on new data distributions \cite{van2022three}. In contrast, humans have an inherent ability to learn new skill over the time without forgetting prior knowledge. In this work, the proposed CELD framework exploits this notion of natural learning ability by retaining the knowledge acquired from previously learned classes to enable the network to adapt to new class. This reduces the requirement for extensive datasets for each incoming new class. This makes it highly suitable for real-world scenarios characterized by limited data availability, such as the classification of retinal fundus images, where obtaining extensive labeled datasets is often a significant challenge. Employing models pre-trained on the ImageNet dataset and subsequently fine-tuning them for adapting to new tasks in the medical domain may lead to suboptimal performance due to the inherent differences in data distribution between natural images and medical images \cite{he2019rethinking}. CELD framework mitigates the issue of performance degradation caused by domain shift since the datasets for source and target tasks belong to the same domain. In this work the source task is to train the classifier for categorizing healthy and DR images and the target task is defined by adapting the classifier from the source task to categorize the input fundus images into DR, Glaucoma and healthy category.

Formally, $\tau_{S}$ and $\tau_{T}$ represent the datasets for source and target task respectively, and $\tau_{S} \subset \tau_{T} \subset \tau$, where $\tau$ represents the universal domain of retinal fundus images. A classifier $C_S$, parameterized by a set of parameters $\omega_S$, is initially trained on source data $ (x_i,y_i) \in \tau_S$. Here $x_i \in \chi_S$ represent the input images and the corresponding labels $y_i \in \mathcal{Y}_S $ where $\mathcal{Y}_S = \{Healthy, DR\}$.

\begin{equation}
    C_{S}: \chi_S \mapsto \mathcal{Y}_S
\end{equation}
\begin{equation}
    \omega_S^* = \arg\min_{\omega_S} \sum_{i=1}^M{\mathcal{L}(\hat{y}_i =  C_{S}(x_i;{\omega_S}), y_i)}
\end{equation}
Here $\mathcal{L}$ is the loss function to be minimized and $M = |\tau_S|$ during training the classifier $C_S$. The optimized weights from the trained classifier $C_{S}$, ${\omega_S^*}$ are then used to initialize a new classifier $C_T$  which classifies $(x_k, y_k) \in \tau_{T}$ where $x_k \in \chi_T$ represent the input images and the corresponding labels $y_k \in \mathcal{Y}_T $ where $\mathcal{Y}_T = \{Healthy, DR, Glaucoma\}$ with $N = |\tau_T|$. Subsequently, $C_T$  is fine-tuned on the extended dataset.

\begin{equation}
    C_{T}: \chi_T \mapsto \mathcal{Y}_T
\end{equation}
\begin{equation}
    \omega_T^* = \arg\min_{\omega_T} \sum_{i=1}^N{\mathcal{L}(\hat{y}_k =  C_{T}(x_k;{\omega_T}), y_k)}
\end{equation}
Here, $\omega_T^*$ is the updated weight of $C_T$ after training on $\tau_T$. The loss function during the incremental learning phase can be defined as: 
\begin{equation}
    \mathcal{L}(\tau_T; \omega_T) = \mathbb{E}_{(x,y) \sim \tau_T} [ - \sum_{c=1}^{C} y_c \log \hat{y}_c]
\end{equation}

where $(x,y) \sim \tau_T$ indicates that the input data $x$ and the corresponding label $y$ is drawn from the expanded dataset $\tau_T$. $y_c$ is the true label for class $c$ and $\hat{y}_c$ is the predicted label. $C= |\mathcal{Y}_T|$ represents the total number of classes in $\tau_T$. The expectation $\mathbb{E}$ denotes averaging over all samples in the dataset. This approach allows $C_T(.)$ to retain patterns learned from $\tau_S$ while learning features relevant to the new class, thus improving the retention of previously learned knowledge and avoiding overfitting.

\subsection{Classifier} 
This paper adapts DenseNet121 \cite{huang2017densely} as the backbone classifier based on the experimental results shown in Sec. \ref{ExpAndRes}. DenseNet121 is characterized by a dense connectivity pattern, where each convolutional layer receives inputs from all preceding layers within a dense block, thereby promoting efficient feature reuse and robust gradient flow. This ensures strong gradient signals even for the earliest layers during backpropagation \cite{huang2017densely}. The architecture consists of 121 convolutional layers, organized into four dense blocks and separated by three transition layers. The transition layers apply normalization, followed by a convolution and pooling operations to downsample the feature maps. Finally, the intermediate feature map obtained undergoes global average pooling and is fed to fully connected layers for classification. The dense connections reduce redundant parameters which lower model complexity and enable faster training of deeper networks \cite{huang2017densely}. This improved information flow helps to reduce overfitting, which is crucial for handling imbalanced data.

\subsection{Perturbation Methods for Explainability}
The black-box nature of deep neural networks hinders the understandability of how predictions are made by the model. This limits the usability of the AI algorithms in critical scenarios like healthcare where the rationale behind the decision-making process of the model must align with the characteristics taken into account by the healthcare professionals. In the realm of deep neural networks, explainability is not just a desirable feature—it is a necessity. Without a clear understanding of how and why these complex models make decisions, particularly in medicine, we risk compromising trust and safety. To make the model more trustworthy and transparent, our framework uses perturbation techniques to identify the relevant characteristics of input data that influence the decision-making process of the model. An efficient model should learn from salient features rather than spurious information or noise that is present in the training data. Perturbation methods, being model-agnostic, allow dynamic analysis without requiring access to the model's internal details. Techniques like applying occlusion masks or adding noise to image patches or pixels help in querying the model and developing test hypotheses on the fly. The main challenge is selecting appropriate perturbation techniques to analyze the model’s performance effectively.

The detection of DR requires identifying pathologies spread across various quadrants of the eye fundus image. In contrast, diagnosing Glaucoma necessitates a precise analysis of the optic disc, focusing on the cup-to-disc ratio. The formation of red and bright lesions are the two most common symptoms of DR. Research shows that the identification of red and bright lesions is most effectively done using the green channel of color fundus images \cite{walter2007automatic}. Additionally, as the condition worsens, neovascularization, which involves the formation of new blood vessels, takes place. Neovascularization in advanced stages of DR can also impact the optic disc area, resulting in the formation of new blood vessels in the optic disc. 

\begin{figure}[!ht]
    \centering
    \includegraphics[width=\linewidth]{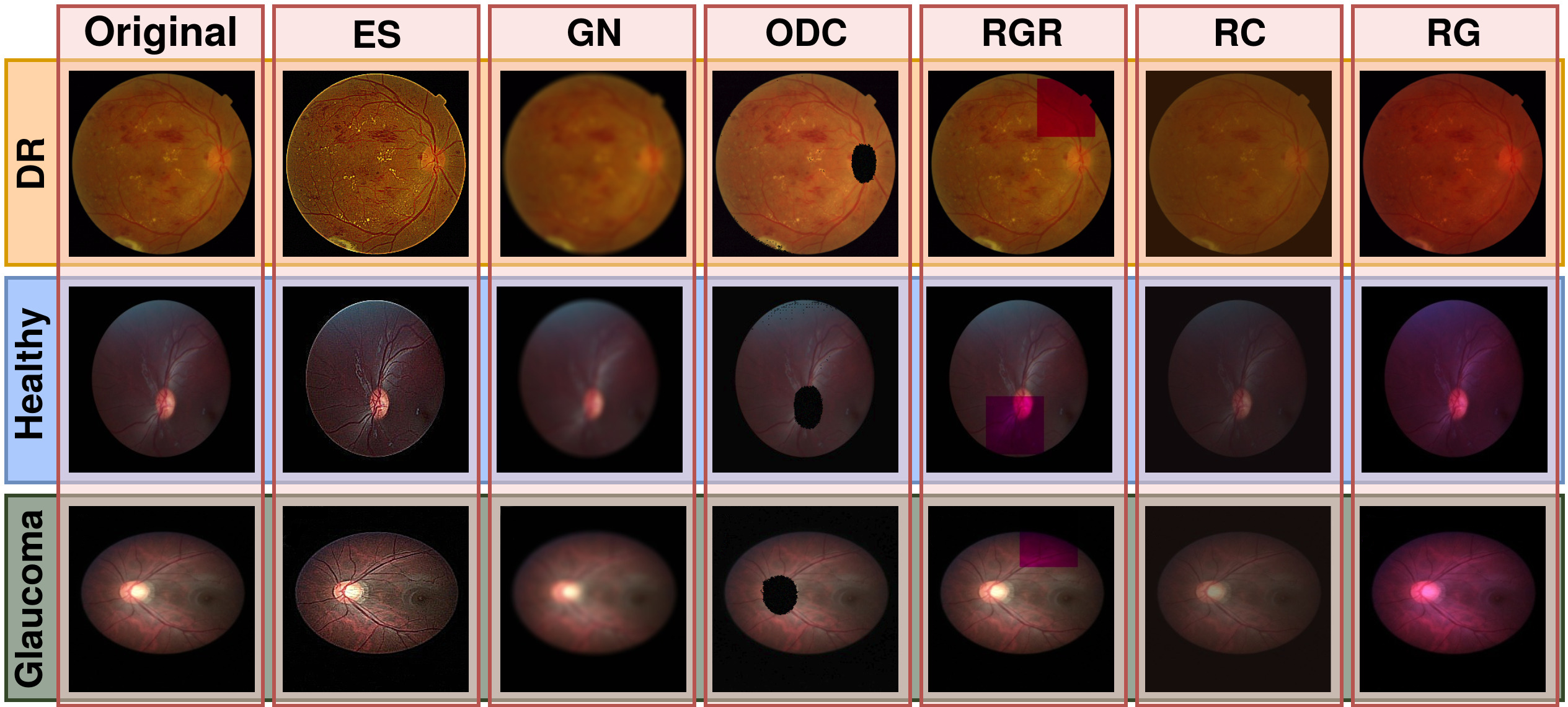}
    \caption{The original image and its perturbated versions for each of the classes: DR, Healthy and Glaucoma.}
    \label{fig:perturbated}
\end{figure}

Based on the above insights gained regarding the relevant clinical features for DR and Glaucoma, we have designed perturbation techniques to further investigate the model's decision-making process. Multiple controlled perturbations are applied to the test dataset and the performance of classifier $C_T$ on this perturbed data was compared with the one obtained from the unperturbed test dataset. Two techniques were used to assess the influence of the green channel in the decision-making process: Reduce green (RG) which reduces the overall green channel weightage in comparison to the red and blue channels of color fundus images and Random green removal (RGR) which randomly removes segments of the green channel. Additional techniques like reducing image contrast (RC), adding Gaussian noise (GN) and applying edge sharpening (ES) were also used to study the impact of image quality on the model's inference. A strategy namely Optic disk occlusion (ODC) was used to evaluate the relevance of the optic disc in the classification of Glaucoma and DR images. Fig. \ref{fig:perturbated} shows the different kind of perturbated images.

\section{Experiments and Results} \label{res}
\label{ExpAndRes}
This section provides a comprehensive overview of the datasets used, the metric used to assess the classifier's performance, the experimental setup, and the resulting experimental outcomes.
\subsection{Dataset}
A total of 3,111 retinal color fundus images were obtained from three publicly available datasets: Messidor2 \footnote{\url{https://www.kaggle.com/datasets/mariaherrerot/messidor2preprocess}}, Chaksu \cite{kumar2023chakṣu}, and LES-AV \cite{orlando2018towards}. The Messidor2 dataset has 1,744 macula-centered RGB images. There are 1017 images belonging to the healthy class and 727 images belonging to the DR category. The Chaksu dataset comprises 1,345 images, with 188 images classified as Glaucoma and 1,157 images classified as healthy. These images were captured using three devices, including two non-mydriatic fundus cameras: the Remido non-mydriatic Fundus-on-Phone (FoP) and the Forus 3Nethra Classic non-mydriatic fundus camera and the Bosch handheld fundus camera. These images are Optic Disc-centered for Optic Disc assessment and Glaucoma detection. The LES-AV dataset has 22 images with 11 images categorized into Glaucoma and the remaining 11 images categorized into healthy category. The data details are given in Table \ref{data}.

\begin{table}[!ht]
\caption{Pooled Dataset Details}
\label{data}
\centering
\begin{tabular}{|p{30mm}|p{15mm}|p{15mm}|p{18mm}|p{15mm}|}
\hline
\textbf{Data} & \textbf{DR}  & \textbf{Healthy}  & \textbf{Glaucoma} & \textbf{Total} \\ \hline
\textbf{Messidor2} & 727 & 1017 & - & 1744 \\ \hline
\textbf{LES-AV} & - & 11 & 11 & 22 \\ \hline
\textbf{Chaksu} & - & 1157 & 188 & 1345 \\ \hline
\textbf{Overall Samples per Class} & 727  & 2185  & 199 & 3111 \\ \hline
\end{tabular}
\end{table}

\textbf{\textit{Data Pooling:}} To address data scarcity, data from three sources were combined to create a more diverse dataset, enabling the model to learn salient features and generalize better. The inclusion of images from different populations and imaging conditions, along with an increased chance of capturing rare medical conditions, reduces model bias toward irrelevant features. After pooling from the three datasets, there are $2185$ healthy, $727$ DR, and $199$ Glaucoma samples. All images were resized to dimensions of $256 \times 256$ from each dataset. The combined dataset was split into $80\%$ for training, $10\%$ for testing, and $10\%$ for validation. To ensure accurate splits, the data was initially divided within each dataset before combining, ensuring that each train-test-validation split contained data from all sources. This approach was applied separately for healthy, DR, and Glaucoma-affected fundus images from each dataset within its split.
\subsection{Experimental Setup}
CELD is developed using Pytorch \footnote{\url{https://pytorch.org/}} and Monai \footnote{\url{https://monai.io/}} on python 3.9 as the platform. All experiments were performed on a 12 GB NVIDIA Titan XP GPU. The initial learning rate was set to $10^{-5}$. Early stopping is used to avoid over-fitting along with the AdamW optimizer\cite{loshchilov2018decoupled}. A batch size of 8 is used in training. 

\subsection{Metric}
The performance of the proposed framework was evaluated using accuracy, precision, recall, and F1-score.  The mathematical definition of the listed metrics in terms of True Positive (TP), False Positive (FP), False Negative (FN) and True Negative (TN) is defined below.

\begin{equation*}
    Accuracy = \frac{TP + TN}{TP + FP + FN + TN}
\end{equation*}

\begin{equation*}
    Precision = \frac{TP}{TP + FP}
\end{equation*}

\begin{equation*}
    Recall = \frac{TP}{TP + FN}
\end{equation*}



\begin{equation*}
    F1-score = 2 \times \frac{Precision \times Recall}{Precsion + Recall}
\end{equation*}

\subsection{Result}
\begin{figure}[!ht]
    \centering
    \includegraphics[width=\linewidth]{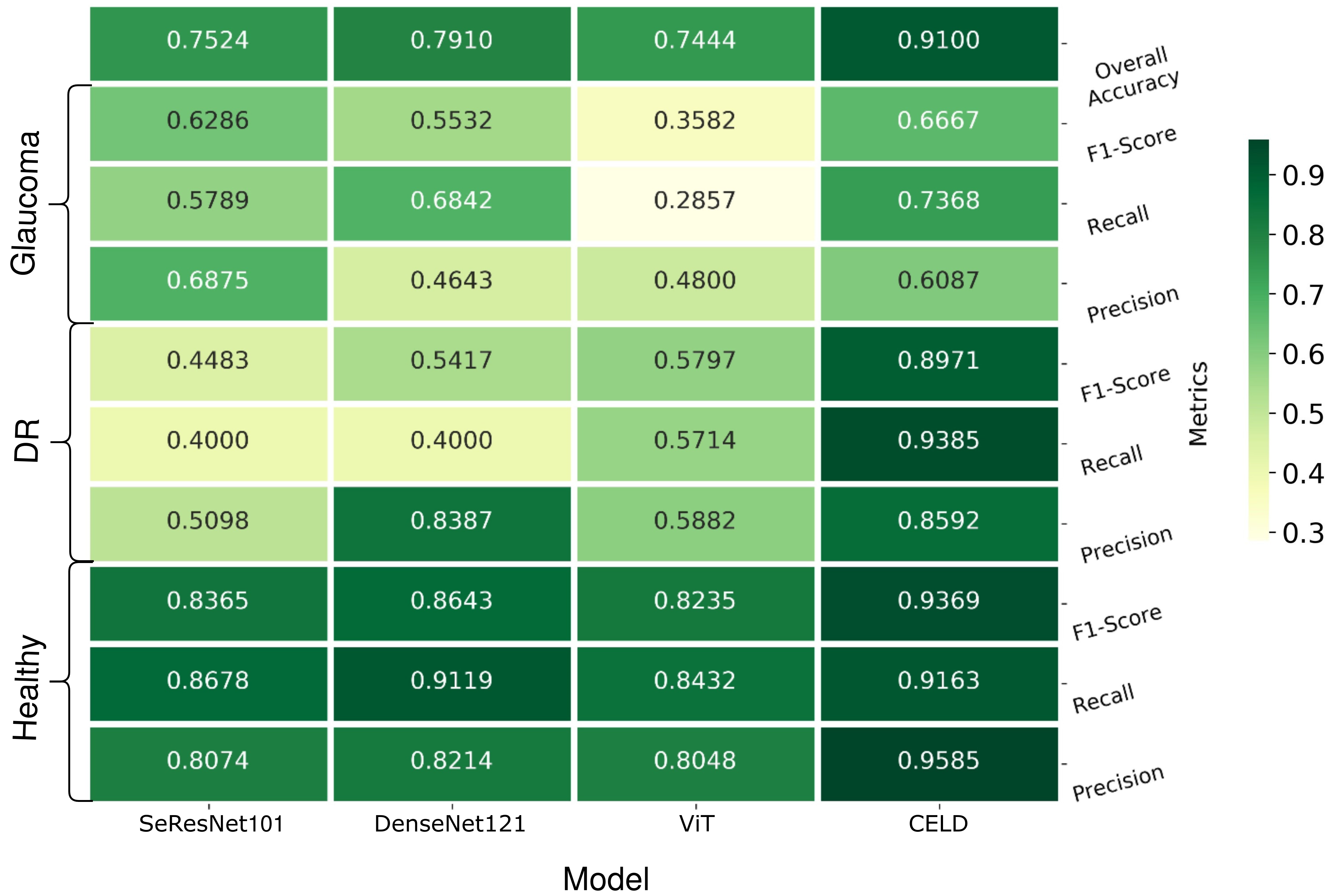}
    \caption{Quantitative Result for 3 Class Classification.}
    \label{fig:Classification3}
\end{figure}

The state-of-the-art (SOTA) models such as SeResNet101 \cite{Hu_2018_CVPR}, DenseNet121, and ViT were employed for classifying retinal images into Healthy, DR and Glaucoma, with their performance summarized in Fig. \ref{fig:Classification3}. DenseNet121 achieved the highest accuracy at $0.7910$. However, all models exhibited poor performance in classifying Glaucoma and DR due to significant class imbalance. 


Further, the same models were tested, with results summarized in Table \ref{Classification2} where the classifiers are trained to categorize images into Healthy and DR only. Significant performance improvements were observed, particularly in the DR class, as indicated by balanced precision-recall scores. DenseNet121 outperformed the other models, achieving an overall accuracy of $0.8729$ and F1-scores of $0.8987$ for healthy and $0.8296$ for DR, leading to its selection as the backbone architecture for the CELD framework.

\begin{table}[!ht]
\caption{Quantitative Result for 2 Class Classfication}
\label{Classification2}
\centering
\scalebox{0.95}{
\begin{tabular}{|l|cccccc|l|}
\hline
\multirow{3}{*}{\textbf{Model}} & \multicolumn{6}{c|}{\textbf{Classes}}                                                                                                                                                                                        & \multirow{3}{*}{\begin{tabular}{c}\textbf{Overall} \\ \textbf{Accuracy} \end{tabular}} \\ \cline{2-7}
                                & \multicolumn{3}{c|}{\textbf{Healthy}}                                                                                    & \multicolumn{3}{c|}{\textbf{DR}}                                                                   &                                            \\ \cline{2-7}
                                & \multicolumn{1}{c|}{\textbf{Precision}} & \multicolumn{1}{c|}{\textbf{Recall}} & \multicolumn{1}{c|}{\textbf{F1-Score}} & \multicolumn{1}{c|}{\textbf{Precision}} & \multicolumn{1}{c|}{\textbf{Recall}} & \textbf{F1-Score} &                                            \\ \hline
\textbf{SeResNet101}              & \multicolumn{1}{c|}{0.8264}             & \multicolumn{1}{c|}{0.8772}          & \multicolumn{1}{c|}{0.8511}            & \multicolumn{1}{c|}{0.7667}             & \multicolumn{1}{c|}{0.6866}          & 0.7244            & \multicolumn{1}{c|}{0.8066}                \\ \hline
\textbf{DenseNet121}            & \multicolumn{1}{c|}{\textbf{0.9027}}    & \multicolumn{1}{c|}{\textbf{0.8947}} & \multicolumn{1}{c|}{\textbf{0.8987}}   & \multicolumn{1}{c|}{\textbf{0.8235}}    & \multicolumn{1}{c|}{\textbf{0.8358}} & \textbf{0.8296}   & \multicolumn{1}{c|}{\textbf{0.8729}}       \\ \hline
\textbf{ViT}                    & \multicolumn{1}{c|}{0.8103}             & \multicolumn{1}{c|}{0.8246}          & \multicolumn{1}{c|}{0.8174}            & \multicolumn{1}{c|}{0.6923}             & \multicolumn{1}{c|}{0.6716}          & 0.6818            & \multicolumn{1}{c|}{0.7680}                 \\ \hline
\end{tabular}
}
\end{table}

For this three-class classification problem, the proposed CELD framework outperformed the state-of-the-art (SOTA) models, achieving an overall accuracy of $0.9100$. The performance of the CELD framework has been listed in Fig. \ref{fig:Classification3}. It demonstrated significant improvement in the F1-scores for all classes, particularly for DR and Glaucoma. While the ViT model achieved a maximum F1-score of $0.5797$ for DR, the CELD framework substantially improved this to $0.8971$, with a high yet balanced precision-recall. Similarly, SeResNet's highest F1-score of $0.6286$ for Glaucoma was improved to $0.6667$ by the CELD framework. In medical image analysis, it is crucial for models to accurately detect positive cases, even if it occasionally results in false alarms. The CELD framework showed significant improvement in recall, albeit with a slight drop in precision for Glaucoma. Overall, the CELD framework significantly outperformed other models across various parameters.

\begin{figure}[!ht]
    \centering
    \includegraphics[width=\linewidth]{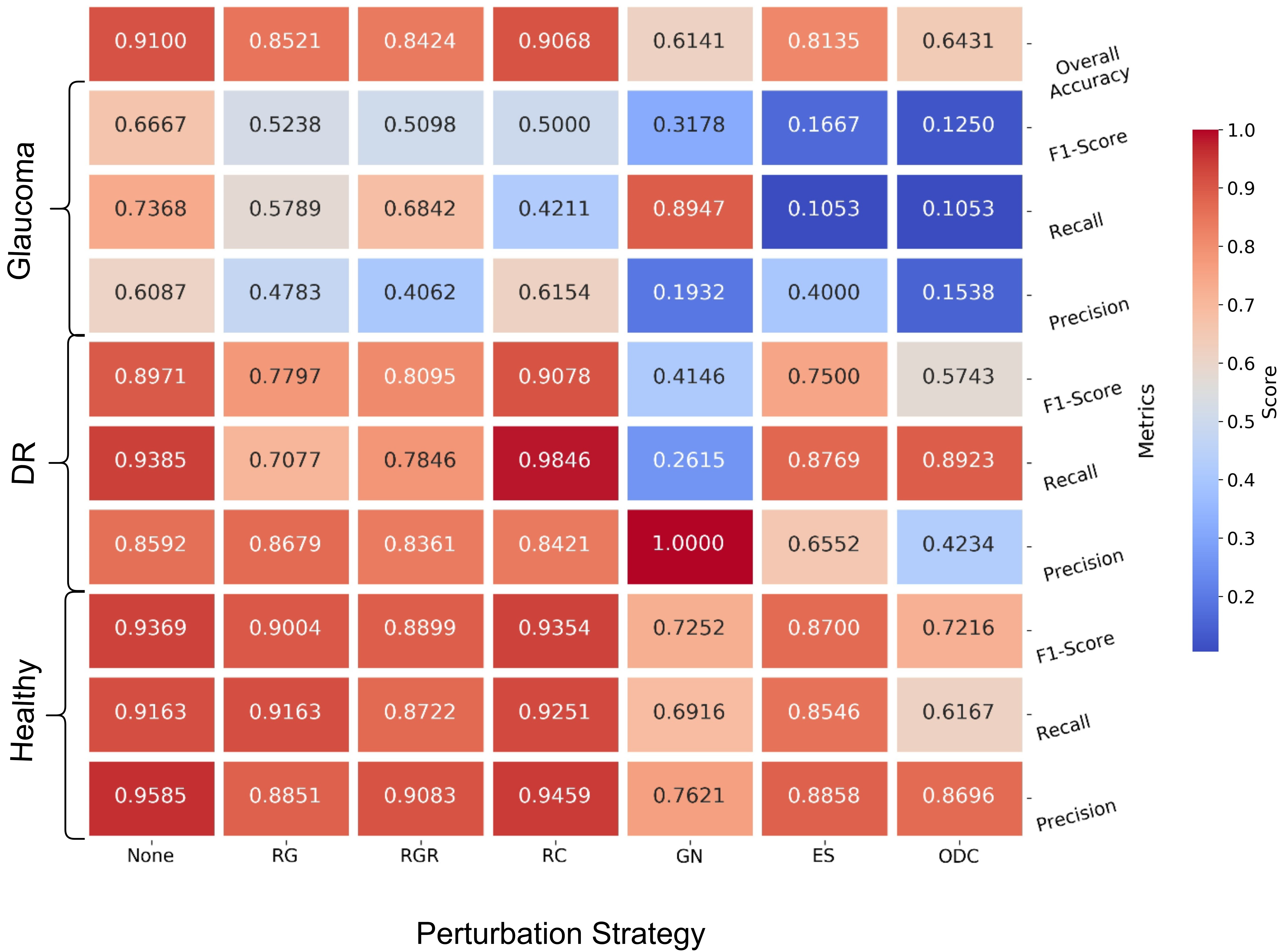}
    \caption{Quantitative Result over CELD framework with Data Perturbation.}
    \label{fig:perturbRes}
\end{figure}

\begin{figure}[!ht]
    \centering
    \includegraphics[width=10cm]{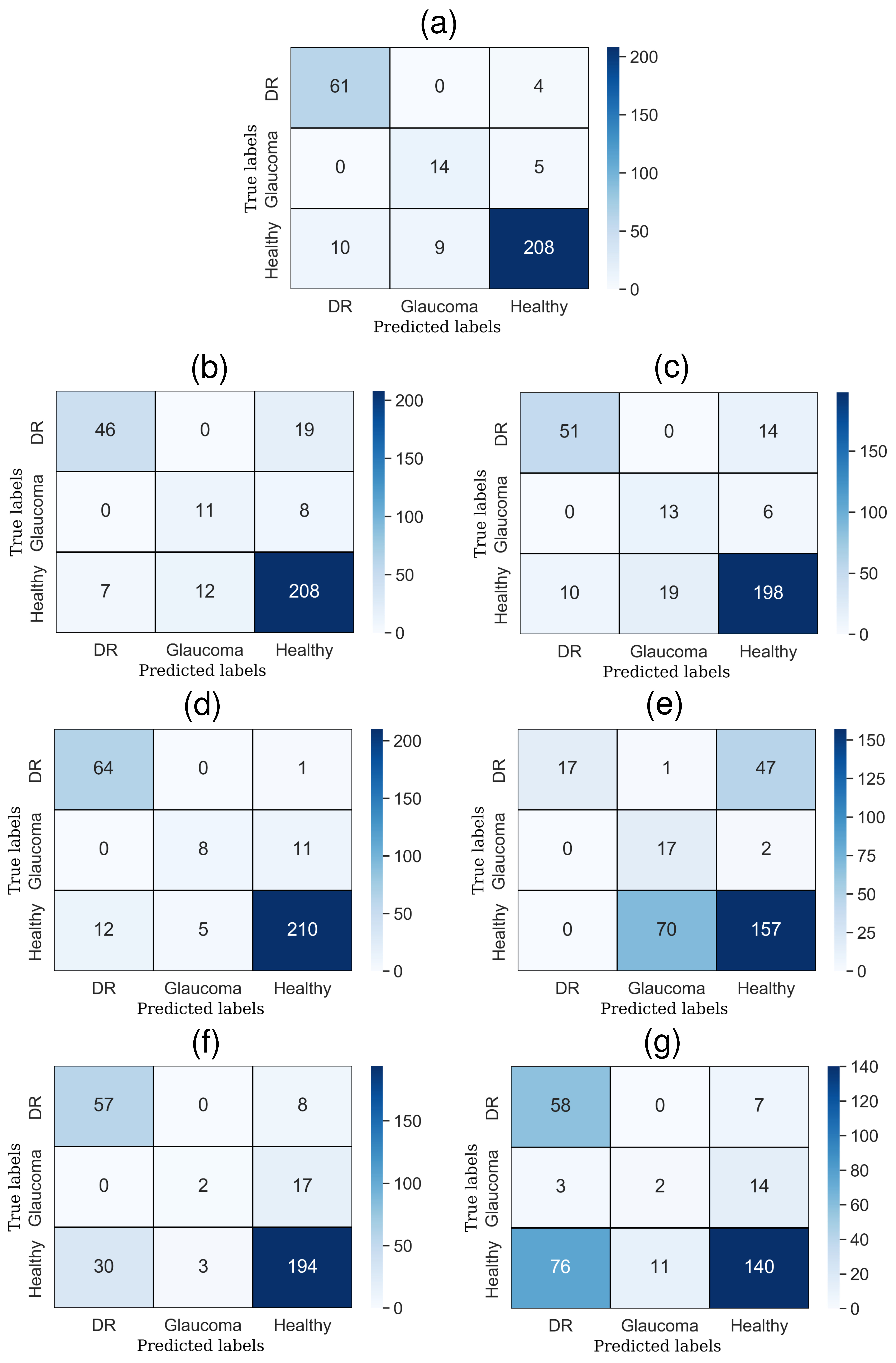}
    \caption{Confusion matrix over CELD framework with Data Perturbation. The matrix shows performance of CELD (a) With no perturbation, (b) Reduce green (RG) , (c) Random green removal (RGR), (d) Reducing image contrast (RC), (e) Gaussian noise (GN), (f) Edge sharpening (ES), (g) Optic disk occlusion (ODC)}
    \label{fig:CM}
\end{figure}

\textbf{\textit{Explaining CELD framework with Data Perturbation: }} 
The performance of the proposed framework was evaluated using perturbed data and compared to unperturbed data, as summarized in Fig. \ref{fig:perturbRes}.The corresponding confusion matrix is represented in Fig. \ref{fig:CM}. Reducing the weight of the green channel significantly decreased performance for DR and Glaucoma classifications, while the classification of healthy samples remained mostly unaffected. The confusion matrix shows increased mis-classification of DR and Glaucoma images as healthy when the green channel’s weight is negatively altered or partially removed. Notably, reducing the green channel's weight across the entire image leads to higher mis-classification rates than randomly removing segments of the green channel. When random patches of green channel are removed, the classifier's decision for the DR class is influenced by the remaining unperturbed data. The performance drop for the Glaucoma class is less significant, as the optic disc region often remains unperturbed in many images.

Reducing contrast does not significantly impact the classification of DR or healthy categories, as shown in the confusion matrix, but it does impair Glaucoma classification. Visually, this perturbation blurs the optic disc region, thus, obscuring key features. Edge sharpening, which enhances pixels having high-intensity changes w.r.t it's neighborhood, leads to a high mis-classification rate of Glaucoma as healthy, as reflected by the F1-score. Adding random Gaussian noise causes the model to falsely classify only one DR image as Glaucoma and most other DR images as healthy, resulting in high precision but a low F1-score for the DR class. It is important to note that in this study this the first perturbation strategy that generates ambiguous decisions between the two disease classes. The model's decision for Glaucoma is less affected by noise but becomes prone to classifying healthy images as Glaucoma, leading to high recall and low precision for the Glaucoma class. In summary, DR identification is challenging in poor-quality images, while Glaucoma can be diagnosed in noisy, low-quality images but not in those with poor contrast or excessive sharpening.

Observing the optic disc occlusion strategy reveals the model's high dependency on the optic disc for classifying Glaucoma and healthy eyes. These two classes exhibit high mis-classification rates, which is understandable since eye experts often diagnose Glaucoma by examining the optic disc region. For Glaucoma, the absence of relevant features leads to mis-classification, as shown in the confusion matrix and Fig. \ref{fig:perturbRes}. Most mis-classified images are labeled as healthy, indicating the model relies on this feature for Glaucoma identification, thereby increasing the precision score for the normal class. The optic disc's features are crucial for determining eye health, reflected in the lower F1-score compared to unperturbed data for healthy eyes. For DR, performance is less affected since neovascularization in the optic disc is not always present in DR-affected images. However, there is an increase in false positives for the DR class due to the model's insufficient features for reliable decisions, resulting in high recall and low precision for the DR class. In conclusion, optic disc occlusion significantly impacts overall model accuracy, highlighting its importance as an input feature.

To conclude, the perturbation techniques revealed that the model heavily relies on the green channel and image quality for accurate classification, especially for DR and Glaucoma. Occlusion of the optic disc significantly impacts Glaucoma detection, emphasizing its critical role in the model's decision-making process.

\section{Conclusion and Discussion} 
\label{concl}

This research demonstrates the potential of deep neural networks to improve medical image classification, particularly for identifying conditions like diabetic retinopathy (DR) and Glaucoma from fundus images. Initially, we trained the network to differentiate between healthy and DR-affected images. Using the Class Extension with Limited Data (CELD) framework, we fine-tuned the model also to classify Glaucoma, transforming it into a three-class classifier. The CELD framework enables the model to maintain its performance on previously learned tasks while adapting to new classes while efficiently tackling data imbalance with minimal computational overhead and data requirements. Consequently, the model retains its proficiency in identifying DR while learning to classify Glaucoma, ensuring efficiency and resource-friendliness.

Our extensive empirical analysis compared the performance of two-class and three-class classifiers. The results highlighted that the Densenet121 architecture significantly improves classification accuracy, proving its suitability for this application. We conducted various experiments to assess accuracy and robustness, confirming the model's effectiveness. Additionally, we explored feature relevance through explainability using perturbed data. These studies provided insights into how changes in input data affect model performance, identifying the most critical features for accurate classification. The perturbation analysis summarized the robustness of the CELD framework. This approach represents a significant advancement in medical imaging and deep learning, providing an efficient method to expand model capabilities with limited data and computational resources. The CELD framework has the potential to be applied to diagnose a variety of other ocular diseases common in diabetic eyes.

%
%
%
\bibliographystyle{splncs04}
\bibliography{reference}

\end{document}